\documentclass[sigplan,10pt]{acmart}

\makeatletter
\def\@ACM@checkaffil{
    \if@ACM@instpresent\else
    \ClassWarningNoLine{\@classname}{No institution present for an affiliation}%
    \fi
    \if@ACM@citypresent\else
    \ClassWarningNoLine{\@classname}{No city present for an affiliation}%
    \fi
    \if@ACM@countrypresent\else
        \ClassWarningNoLine{\@classname}{No country present for an affiliation}%
    \fi
}
\makeatother

\usepackage{graphicx}
\usepackage{pifont}
\usepackage{epstopdf}
\usepackage{subcaption}
\usepackage[misc]{ifsym}
\usepackage{breakurl}
\usepackage{multirow}
\usepackage{hyperref}
\usepackage{pifont}
\usepackage{breakurl}
\usepackage{url}
\usepackage{geometry}

\AtBeginDocument{%
  }

\copyrightyear{2024} 
\acmYear{2024} 
\setcopyright{acmlicensed}\acmConference[SOSP '24]{ACM SIGOPS 30th Symposium on Operating Systems Principles}{November 4--6, 2024}{Texas, USA}
\acmBooktitle{ACM SIGOPS 30th Symposium on Operating Systems Principles (SOSP '24), November 4--6, 2024, Texas, USA}

\copyrightyear{2024}
\acmYear{2024}
\setcopyright{acmlicensed}\acmConference[SOSP '24]{ACM SIGOPS 30th
Symposium on Operating Systems Principles}{November 4--6, 2024}{Austin,
TX, USA}
\acmBooktitle{ACM SIGOPS 30th Symposium on Operating Systems Principles
(SOSP '24), November 4--6, 2024, Austin, TX, USA}
\acmDOI{10.1145/3694715.3695964}
\acmISBN{979-8-4007-1251-7/24/11}

\usepackage{titlesec}
\titlespacing*\section{3pt}{3pt plus 1pt minus 1pt}{3pt plus 1pt minus 1pt}
\titlespacing*\subsection{2pt}{2pt plus 1pt minus 1pt}{2pt plus 1pt minus 1pt}
\begin{document}

\newcommand{\sys}{PowerInfer}

\newcommand{\TODO}[1]{\textcolor{red}{TODO: #1}}

\title{{\sys}: Fast Large Language Model Serving with a Consumer-grade GPU}


\author{\rm Yixin Song, Zeyu Mi, Haotong Xie and Haibo Chen \\
{\normalsize \it {Institute of Parallel and Distributed Systems, SEIEE, Shanghai Jiao Tong University}} \\
} 
\renewcommand{\authors}{%
Yixin Song, Zeyu Mi, Haotong Xie and Haibo Chen}
\renewcommand{\shortauthors}{%
Yixin Song, Zeyu Mi, Haotong Xie and Haibo Chen}

\sloppy

\begin{abstract}

This paper introduces {\sys}, a high-speed Large Language Model (LLM) inference engine on a personal computer (PC) 
equipped with a single consumer-grade GPU. The key principle underlying the design of {\sys} is exploiting the high locality 
inherent in LLM inference, characterized by a power-law distribution in neuron activation. 
This distribution indicates that a small subset of neurons, termed \textit{hot neurons}, are consistently activated 
across inputs, while the majority, \textit{cold neurons}, vary based on specific inputs.
{\sys} exploits such an insight to design a GPU-CPU hybrid inference engine:
hot-activated neurons are preloaded onto the GPU for fast access, while cold-activated neurons are computed 
on the CPU, thus significantly reducing GPU memory demands and CPU-GPU data transfers.
{\sys} further integrates adaptive predictors and neuron-aware sparse operators,
optimizing the efficiency of neuron activation and computational sparsity.
The evaluation shows that {\sys} significantly outperforms \emph{llama.cpp} by up to 11.69$\times$ while retaining model accuracy across various LLMs (including OPT-175B) on a single NVIDIA RTX 4090 GPU.
For the OPT-30B model, {\sys} achieves performance comparable to that of a high-end server-grade A100 GPU,
reaching 82\% of its token generation rate on a single consumer-grade RTX 4090 GPU.

\end{abstract}




\settopmatter{printfolios=false,printacmref=false}

\maketitle

\renewcommand*{\thefootnote}{{\ding{41}}}
\renewcommand{\thefootnote}{\arabic{footnote}}
\setcounter{footnote}{0}





\vspace{-2mm}
\section{Introduction}
\label{sec:intro}

Generative large language models (LLMs) have garnered significant attention for their remarkable capabilities in sophisticated natural language processing tasks~\cite{brown2020language, zhang2022opt, touvron2023llama},
especially in areas such as creative writing and advanced code generation.
These models, widely deployed in data centers equipped with high-end and expensive server-grade GPUs,
have significantly influenced our daily lives and work practices.
Meanwhile, there is an emerging trend of running LLMs on more accessible local platforms~\cite{xue2024powerinfer, 10.1145/1966445.1966473}, particularly personal computers (PCs) with consumer-grade GPUs.
This evolution is driven by the need for enhanced data privacy~\cite{Martinez_Toro_PrivateGPT_2023},
model customization~\cite{lyu2023llmrec}, and reduced inference costs~\cite{touvron2023llama}.
In contrast to data-center deployments, which prioritize high throughput~\cite{kwon2023efficient, sheng2023flexgen, yu2022orca},
local deployments focus on low latency in processing small batches.

Nonetheless, deploying LLMs on consumer-grade GPUs presents significant challenges due to their substantial memory requirements.
LLMs, functioning as autoregressive Transformers,
sequentially generate text token-by-token,
each needing to access the entire model containing hundreds of billions of parameters.
Therefore, the inference process is fundamentally constrained by the GPU’s memory capacity.
This limitation is particularly acute in local deployments where the processing of individual requests (often one at a time)~\cite{medusa} leaves minimal opportunity for parallel processing.

Compression techniques like quantization~\cite{frantar-gptq, xiao2023smoothquant} and pruning~\cite{ma2023llm}
can reduce the model size.
However, even deeply compressed models remain too large for consumer-grade GPUs.
For instance, an OPT-66B model with 4-bit precision demands approximately 40GB of memory just to load its parameters~\cite{liu2023emergent},
exceeding the capacity of even high-end GPUs like the NVIDIA RTX 4090.

Model offloading is another approach that partitions the model between GPU and CPU at the Transformer layer level~\cite{aminabadi2022deepspeed, sheng2023flexgen, llama.cpp}.
State-of-the-art systems like llama.cpp~\cite{llama.cpp} distribute layers between CPU and GPU memories, leveraging both for inference, thus reducing the GPU resources required.
However, this method is hindered by the slow PCIe interconnect and the CPUs' limited computational capabilities, resulting in high inference latency.

In this paper, we argue that the key reason for memory issues in LLM inference is the \textbf{locality mismatch} between
hardware architecture and the characteristics of LLM inference.
Current hardware architectures are designed with a memory hierarchy optimized for data locality.
Ideally, a small, frequently accessed working set should be stored in the GPU, which offers higher memory bandwidth but limited capacity.
In contrast, larger, less frequently accessed data are better suited for CPUs,
which provide more extensive memory capacity but lower bandwidth.
Nevertheless, each LLM inference iteration requires accessing the entire set of model parameters whose total size is too large for a single GPU,
thus showing no locality at all and thus impeding efficient locality exploitation.

Recent works have identified activation sparsity in LLM inference~\cite{liu2023deja, mirzadeh2023relu, zhang2024relu2}.
During each inference iteration, only a limited number of neurons\footnote{This paper defines a neuron as a specific row/column in a weight matrix.} are activated,
significantly influencing token outputs.
These sparse activations, which can be accurately predicted at runtime,
allow for accelerated inference by computing only the activated neurons.
However, the set of activated neurons varies across inputs and can only be determined at runtime,
necessitating the entire model to be loaded into GPU memory.
This requirement limits the approach's applicability in local deployment scenarios with constrained GPU VRAM.

Fortunately, we have observed that neuron activation in an LLM follows a \textbf{skewed power-law distribution} across numerous inference processes:
a small subset of neurons \textit{consistently} contribute to the majority of activations (over 80\%)
across various inputs (hot-activated),
while the majority are involved in the remaining activations, which are determined based on the inputs at runtime (cold-activated).
This observation suggests an inherent locality in LLMs with high activation sparsity, which could be leveraged to address the aforementioned locality mismatch.

Building on this locality insight,
we introduce \textbf{\sys}, an efficient LLM inference system optimized for local deployments using a single consumer-grade GPU.
The key idea of {\sys} is to exploit the locality in LLM inference by assigning the minor hot neurons to the GPU,
while cold neurons, which constitute the majority, are managed by the CPU.
Specifically, {\sys} exploits the locality in LLM inference through a two-step process:
(1) {\sys} preselects hot and cold neurons based on their statistical activation frequency, preloading them onto the GPU and CPU, respectively, during an offline phase.
(2) At runtime, it employs online predictors to identify which neurons (both hot and cold) are likely to be activated for each specific input.
This approach allows the GPU and CPU to independently process their respective sets of activated neurons,
thereby minimizing the need for costly PCIe data transfers.

However, there are significant challenges that complicate the design of {\sys}.
First, the online predictors, which are essential for identifying active neurons in LLM layers and are typically situated on the GPU,
occupy a considerable amount of GPU memory.
This memory could otherwise be used for the LLM.
To address this, {\sys} introduces an adaptive method for constructing smaller predictors for layers with higher activation sparsity and skewness.
This iterative process reduces the size of the predictors while maintaining their accuracy, thus freeing up GPU memory for LLM inferences.

Second, leveraging LLM sparsity requires the use of sparse operators.
Conventional libraries like cuSPARSE~\cite{cuSPARSE} are not optimal due to their general-purpose design,
which includes tracking each non-zero element and converting dense matrices into sparse formats~\cite{xia2023flashllm, zheng2023pit}.
In contrast, {\sys} designs neuron-aware sparse operators that directly interact with individual neurons,
thereby bypassing operations on entire matrices.
This approach enables efficient matrix-vector multiplication at the neuron level and removes the need for specific sparse format conversions.

Lastly, the optimal placement of activated neurons between the GPU and CPU in {\sys} is a complex task.
It involves evaluating each neuron's activation rate, intra-layer communication, and available hardware resources like GPU memory sizes.
To effectively manage this, {\sys} utilizes an offline phase to generate a neuron placement policy.
This policy uses a metric that measures each neuron's impact on LLM inference outcomes and is framed as an integer linear programming problem.
The policy formulation considers factors such as neuron activation frequencies and the bandwidth hierarchy of CPU and GPU architectures.

The online inference engine of {\sys} was implemented by extending llama.cpp with an additional 4,200 lines of C++ and CUDA code.
Its offline component, comprising a profiler and a solver, builds upon the transformers framework~\cite{wolf-etal-2020-transformers}
with approximately 400 lines of Python code.
{\sys} is compatible with various popular LLM families, including OPT (7B-175B), LLaMA2 (7B-70B), and Falcon-40B,
and supports consumer-grade GPUs like the NVIDIA RTX 4090 and NVIDIA RTX 2080Ti.

Performance evaluation reveals that {\sys}, when deployed on a PC equipped with a single NVIDIA RTX 4090 GPU,
delivers an average generation speed of 13.20 tokens/s for quantized models and 8.32 tokens/s for non-quantized models, maintaining model accuracy.
These results significantly surpass llama.cpp's performance, exhibiting up to 8.00× and 11.69× improvements for quantized and non-quantized models, respectively.
Significantly, the inference speed achieved on an NVIDIA RTX 4090 GPU (priced at approximately \$2,000)
is only 18\% slower compared to the performance on a top-tier A100 GPU (costing around \$20,000) that can fully accommodate the model.
{\sys}'s code has been open sourced completely.

\vspace{-2mm}
\section{Background and Motivation}
\label{sec:bg}

\subsection{LLM Inference \& Architecture}

LLM inference, an autoregressive model, generates each token based on previous ones.
The process 
starts with a prompt and unfolds in two phases:
first, the prompt phase outputs an initial token,
then the generation phase sequentially produces tokens until a maximum limit or an end-of-sequence (<EOS>) token is reached.
Each token generation, an inference iteration, requires running the full LLM model.

\begin{figure}[htp]
    \vspace{-4mm}
    \centering 
    \setlength{\abovecaptionskip}{0pt}
    \setlength{\belowcaptionskip}{0pt}
    \includegraphics[scale=1.8]{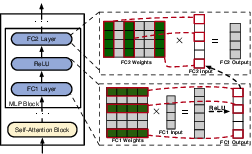}
    \caption{The architecture of a Transformer layer and how neurons are sparsely activated in FC1 and FC2 layers due to the ReLU function.
    The neurons that are activated are represented as green rows or columns encircled by red lines.
    The output vector from FC1 is then supplied to FC2 as its input vector.} 
    \label{fig:bg-mlp}
    \vspace{-3mm}
\end{figure}

The LLM architecture includes multiple Transformer layers,
each comprising a self-attention and an MLP (Multi-Layer Perceptron) block (see Figure~\ref{fig:bg-mlp}, left).
The self-attention block generates embedding vectors by capturing the relationships among input tokens.
In this process, different heads focus on extracting distinct feature information.
The computation results from these different heads are aggregated and then utilized as the input for the MLP block.
The MLP block applies non-linear transformations via fully connected layers and activation functions to refine the input sequence representation.
The output either advances to subsequent layers or forms the LLM's final output.

In Figure~\ref{fig:bg-mlp} (right),
the MLP block's layers, FC1 and FC2, generate vectors through matrix multiplication.
Each output element comes from the dot product of an input vector and a neuron (a row/column in a weight matrix).
Activation functions (like ReLU~\cite{agarap2018deep} and SiLU~\cite{ramachandran2017searching}) act as gates to selectively retain or discard values in a vector, influencing neuron activations in FC1 and FC2.
For example, ReLU in this figure filters out negative values,
allowing only positively valued neurons in FC1 to influence the output.
These neurons, which contribute to the output, are considered activated in this paper.
Similarly, these values also affect which neurons in FC2 are activated and involved in the computation of its output.

\textbf{Activation Sparsity.}
LLM inference exhibits notable sparsity in neuron activation, a phenomenon observed in both self-attention and MLP blocks~\cite{li2022lazy,liu2023deja,zhang2022moefication}.
In self-attention, nearly half of the attention heads contribute minimally, while in MLP blocks, sparsity is largely due to activation function characteristics.
This MLP sparsity is particularly pronounced in ReLU-based architectures, with DejaVu~\cite{liu2023deja} reporting that approximately 80\% of neurons in the OPT-30B model remain inactive during inference.
Moreover, this phenomenon is not limited to ReLU-based models; it is also observed in architectures using other activation functions such as SwiGLU.
For instance, Table~\ref{tab:model_sparsity} shows that LLaMA2-13B and Yi-34B, which employ SwiGLU, exhibit sparsity levels of 43\% and 53\% respectively.
These findings align with prior research results from studies like CATS~\cite{lee2024cats} and ReLU2~\cite{zhang2024relu2}.

\begin{table}[!t]
    \caption{Average activation sparsity for various LLMs in the MLP blocks.
    For ReLU-based models, sparsity is the proportion of neurons with zero activation.
    For SwiGLU-based models, it's the proportion of neurons that can be dynamically pruned with less than 1\% impact on perplexity.}
    \label{tab:model_sparsity}
    \vspace{-3mm}
    \scriptsize{
    \resizebox{0.99\linewidth}{!}{
    \begin{tabular}{ccc}
    \toprule
    \textbf{LLM} & \textbf{Activation Function} & \textbf{Sparsity} \\ \midrule
    OPT-30B & ReLU & 97\%  \\
    LLaMA2-13B & SwiGLU & 43\%  \\
    Yi-34B & SwiGLU & 53\%  \\
    \toprule
    \end{tabular}
    }
    }
\end{table}

Moreover, it is possible to predict neuron activations a few layers in advance within the ongoing model iteration.
Based on this observation, DejaVu~\cite{liu2023deja} employs MLP-based online predictors during inference and only processes the activated neurons,
achieving over a 6x speedup while maintaining an impressive accuracy rate of at least 93\% in predicting neuron activation.
However, the activation sparsity is input-specific for each inference iteration,
meaning that the activation of specific neurons is directly influenced by the current input and cannot be predetermined before the model's inference iteration begins.

\subsection{Offloading-based LLM Serving}
\label{subsec:bg-offloading}

\begin{figure}[t]
    \begin{minipage}{1\linewidth}
        \centering\includegraphics[scale=0.52]{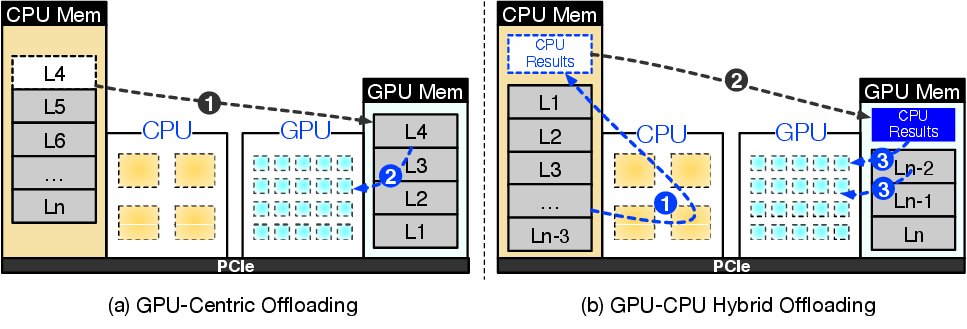}
    \end{minipage}
    \begin{minipage}{1\linewidth}
    \vspace{-3mm}
    \caption{Typical existing offloading solutions. (a) shows a GPU-centric approach, while (b) is the CPU-GPU hybrid offloading approach.}
    \label{fig:existing-arch}
    \end{minipage} 
    \vspace{-5mm}
\end{figure}

\begin{figure}[ht]
    \subfloat[] {
        \includegraphics[width=0.45\linewidth]{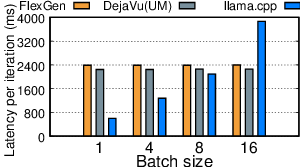}
        \label{fig:eval-latency}
    }
    \subfloat[] {
        \includegraphics[height=2.25cm]{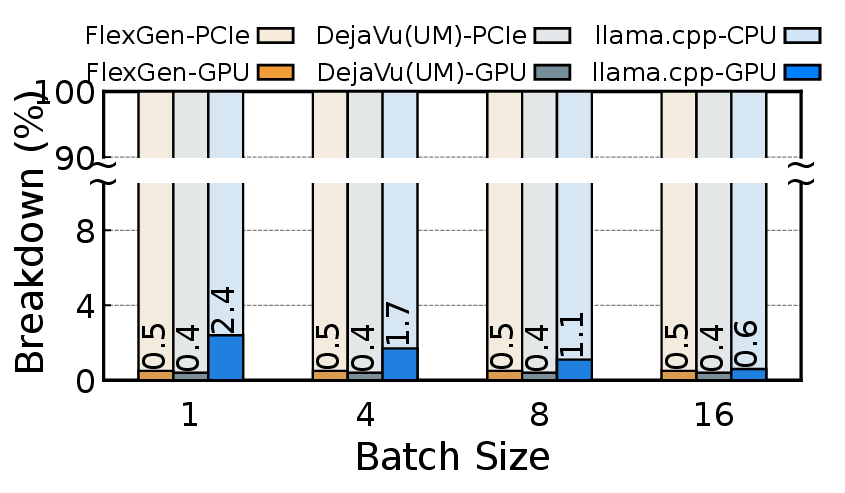}
        \label{fig:eval-breakdown-ratio}
    }
    \vspace{-3mm}
    \caption{\small{\emph{{Performance comparison and analysis for serving OPT-30B on NVIDIA RTX 4090 GPU.}} The \textit{yellow} blocks refer to FlexGen, the \textit{gray} blocks refer to DejaVu (UM) and the \textit{blue} blocks refer to llama.cpp. (a) The Y-axis indicates execution time for one iteration and the X-axis represents batch sizes for input. (b) The Y-axis indicates the proportion of execution time, and the X-axis indicates batch sizes for input.}}
\end{figure}

The offloading technique, which leverages the CPU's additional computational and memory resources, presents a more viable solution for accommodating LLMs that exceed the GPU's memory capacity.
In this section, we delve into the analysis of offloading systems to uncover the factors contributing to their sluggish performance.
Figure~\ref{fig:existing-arch} illustrates two main offloading approaches:

\textbf{GPU-centric offloading} utilizes CPU memory to store portions of the model parameters that exceed the GPU's capacity.
During each iteration, as depicted in Figure~\ref{fig:existing-arch}a), it processes the parameters located in the GPU memory,
transferring more from the CPU as needed.
This strategy enables the inference of LLMs of varying sizes,
provided that sufficient combined CPU memory and hard disk storage are available.
FlexGen~\cite{sheng2023flexgen} is a typical example that adopts a zig-zag scheduling approach to prioritize throughput over latency,
processing batches sequentially for each layer.
Nonetheless, this method leads to substantial per-token latency in latency-sensitive scenarios (Figure~\ref{fig:eval-latency}),
mainly due to frequent data transfers between GPU and CPU.
Over 99.5\% of processing time is consumed by transferring LLM weights from CPU to GPU, significantly impacting overall latency,
as illustrated in Figure~\ref{fig:eval-breakdown-ratio}.

Although DejaVu~\cite{liu2023deja} leverages activation sparsity to accelerate LLM inference,
this approach, originally designed for data center environments, faces challenges when applied to consumer-grade GPUs incapable of hosting full-scale LLMs.
The key challenge with DejaVu in such contexts stems from the need to frequently transfer activated neurons from the CPU to the GPU during runtime.
For LLMs like OPT-30B that exceed GPU memory limits,
DejaVu\footnote{Since DejaVu only works for GPU, we modified it by using NVIDIA Unified Memory (UM)~\cite{UM} to fetch parameters from CPU memory.}, albeit reducing the computational load on the GPU,
is constrained by the data transfer procedure (Figure~\ref{fig:eval-latency}).
Consequently, as shown in Figure~\ref{fig:eval-latency}, DejaVu experiences significant inference latency, comparable to that of FlexGen.

\textbf{Hybrid offloading} distributes model parameters between GPU and CPU, splitting them at the Transformer layer level (Figure~\ref{fig:existing-arch}b), with llama.cpp~\cite{llama.cpp} as an example.
The CPU processes its layers first, then sends intermediate results to the GPU for token generation.
This offloading method reduces inference latency to around 600ms (Figure~\ref{fig:eval-latency}) by minimizing data transfer and mitigating slow PCIe bandwidth. 
However, compared to the 45ms latency of the 30B model on A100, the speed is still too slow.

Hybrid offloading still faces the locality mismatch issue, leading to suboptimal latency.
Each inference iteration accesses the entire model, resulting in poor locality for hierarchical GPU-CPU memory structures.
GPUs, while computationally powerful, are constrained by memory capacity.
For instance, a 30B-parameter model on a 24GB NVIDIA RTX 4090 GPU means only 37\% of the model is on the GPU,
shifting most computational tasks to the CPU.
The CPU, with higher memory but lower computational power, ends up handling 98\% of the total computational time(Figure~\ref{fig:eval-breakdown-ratio}).
\section{Insights into Locality in LLM Inference}
\label{sec:observation}

\subsection{Insight-1: Power-law Activation}
\begin{figure}[ht]
    \subfloat[] {
    \includegraphics[width=0.48\linewidth]{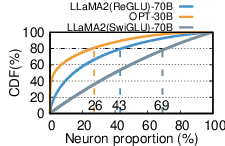}
    }
    \subfloat[] {
        \includegraphics[width=0.48\linewidth]{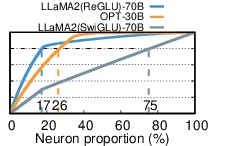}
    }
    \vspace{-3mm}
    \caption{\small{Cumulative distribution function (CDF) of neuron activation in OPT-30B, LLaMA2(ReGLU)-70B and LLaMA2(SwiGLU)-70B.
    (a) CDF in a single MLP block. (b) CDF across the entire model.
    The X-axis shows neuron proportion. The Y-axis represents the CDF of neuron activation.}}
    \label{fig:CDF-powerlaw}
    \vspace{-3mm}
\end{figure}

LLM inference exhibits a high degree of locality,
indicating that a consistent group of neurons is \emph{frequently} activated.
Notwithstanding the input dependence of LLM activation sparsity,
a power-law distribution is evident among activated neurons.
We profile from wikipedia~\cite{wikidump} dataset to collect the statistics of the activation sparsity about 1M tokens.
Figure~\ref{fig:CDF-powerlaw}a reveals that in the MLP blocks of OPT-30B (using ReLU), LLaMA2 (ReGLU)-70B and LLaMA2 (SwiGLU)-70B,
26\%, 43\% and 69\% of neurons respectively are responsible for 80\% of total activations.
This indicates that these neurons are frequently activated, which we termed as \textit{hot-activated} neurons.
Conversely, the activation of the remaining 74\%, 57\% and 31\% of neurons is input-dependent,
classifying them as \textit{cold-activated} neurons. 
This distribution is particularly pronounced in ReLU-based models compared to those using other activation functions due to its higher sparsity.

\begin{figure}[ht]
    \includegraphics[width=0.95\linewidth]{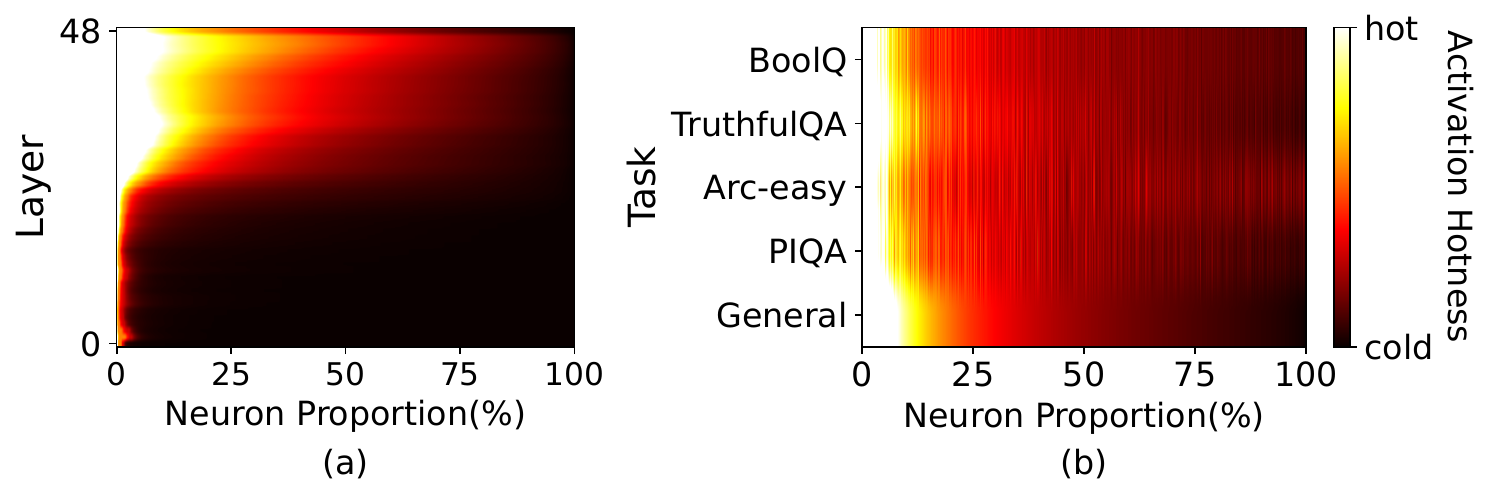}
    \vspace{-5mm}
    \caption{\small{ Activation frequency in OPT-30B.
    (a) The activation frequency of neurons in different layers. The Y-axis represents the layer id. (b) The activation frequency of neurons in different tasks for 30th layer. The Y-axis represents the tasks.
    The X-axis shows neuron proportion.}}
    \label{fig:hot}
    \vspace{-3mm}
\end{figure}

Furthermore, this high locality is not confined to a single MLP block but extends throughout the entire model.
As illustrated in Figure~\ref{fig:CDF-powerlaw}b,
approximately 17\% of neurons in OPT-30B, 26\% in LLaMA2 (ReGLU)-70B, and 75\% in LLaMA2 (SwiGLU)-70B are responsible for 80\% of the total activations across all layers.
Figure~\ref{fig:hot}a clearly demonstrates that in the OPT-30B model,
each transformer layer contains a small portion of hot neurons, while the majority are cold neurons. 
Notably, in the initial 24 layers, OPT-30B exhibits exceptionally low activation,
with less than 1\% of neurons becoming active, resulting in a minimal presence of hot neurons within these layers.

Interestingly, the same group of hot neurons consistently remains active across various downstream tasks.
To demonstrate this, we profiled neuron activation across diverse datasets, including knowledge-based, truthfulness assessment, and reasoning tasks.
Figure~\ref{fig:hot}b illustrates that these hot neurons are consistently activated consistently across different tasks,
indicating a stable pattern of power-law activation.
Our analysis revealed that there is over 90\% overlap in the top 20\% most frequently activated neurons across these varied domains.
This remarkable consistency suggests that the power-law distribution of neuron activations is an intrinsic property of the model architecture rather than being dataset-specific.

\subsection{Insight-2: Fast In-CPU Computation}

\begin{figure}[ht]
    \subfloat[MLP block] {
        \includegraphics[width=0.47\linewidth]{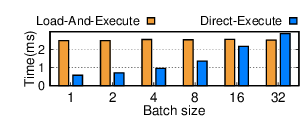}
    }
    \subfloat[Attention block] {
        \includegraphics[width=0.47\linewidth]{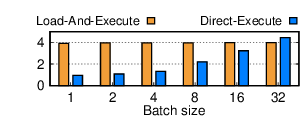}
    }
    \vspace{-3mm}
    \caption{\small{Comparison of execution time for load-then-execute versus direct-execute methods when 10\% and 60\% neuron weights of one MLP and attention block in OPT-30B are CPU-resident.
    The X-axis shows input batch sizes, and the Y-axis measures execution time (ms).
    Load-then-execute involves transferring these neuron weights to GPU memory for computation, whereas direct-execute computes them directly on the CPU.
    }}
    \label{fig:CPU_vs_GPU_calculate}
    \vspace{-2mm}
\end{figure}

If activated neurons reside in CPU memory, computing them on the CPU is faster than transferring them to the GPU,
especially with the small number of activated neurons and the small batch sizes typical in local deployments.
Modern CPUs with vector extensions can efficiently handle such smaller matrix computations.

We compared the time to load and compute 10\%\footnote{While Insight-1 indicates that 43\% of neurons account for 80\% of the total activations in a single MLP block, it is typically found that only about 10\% of its neurons are activated during an individual inference iteration.} sparisty
of the MLP block and 60\% of attention block's CPU-side neurons on the GPU versus direct CPU execution in OPT-30B.
Results in Figure~\ref{fig:CPU_vs_GPU_calculate} indicate that for batch sizes under 32,
the time taken to transfer the weights of these neurons and compute them on the GPU (NVIDIA RTX 4090)
exceeds the time required for calculation directly on the CPU using the AVX2 vector extension.
\vspace{-1mm}
\section{{\sys} Overview}
\label{sec:overview}

We present {\sys}, a low-latency LLM inference system deployed in a PC equipped with a single consumer-grade GPU.
{\sys} proposes a neuron-aware offloading strategy and an inference engine by fully leveraging the high locality insights described in $\S$\ref{sec:observation}.
It utilizes both GPU and CPU for weight storage,
accommodating LLMs of various sizes.
This offloading approach, based on \textit{Insight-1}, effectively exploits the power-law distribution of LLM inference.
Specifically, {\sys} preloads the GPU with weights for neurons that activate frequently, while less active neurons' weights are kept on the CPU.

To reduce inference latency,
the inference engine computes only neurons predicted as active by online predictors,
skipping most inactive ones.
Moreover,
the preloading strategy enables {\sys} to allocate the bulk of inference tasks to the GPU,
given that hot-activated neurons that have been loaded on the GPU constitute a major fraction of activations.
For cold-activated neurons not in GPU memory,
{\sys} executes their computations on the CPU, eliminating the need for weight transfers to the GPU (\textit{Insight-2}).

The effectiveness of {\sys} is directly correlated with the model's activation sparsity.
ReLU-family LLMs, exhibiting over 90\% sparse activations in their FFNs, are ideal candidates for {\sys}.
While LLMs with other activation functions typically show around 50\% sparsity, resulting in less pronounced acceleration,
{\sys} still offers some performance gains.
Encouragingly, there is a growing trend on enhancing LLM sparsity without compromising performance~\cite{song2024turbo,song2024prosparse},
or directly training LLMs with ReLU-family activations, as seen in NVIDIA Nemotron~\cite{adler2024nemotron} and MiniTron~\cite{sreenivas2024llm}.
This trend is expected to extend {\sys}'s applicability across a more diverse spectrum of LLMs.

\subsection{Architecture and Workflow}
\begin{figure}[t]
    \begin{minipage}{1\linewidth}
        \centering\includegraphics[scale=0.75]{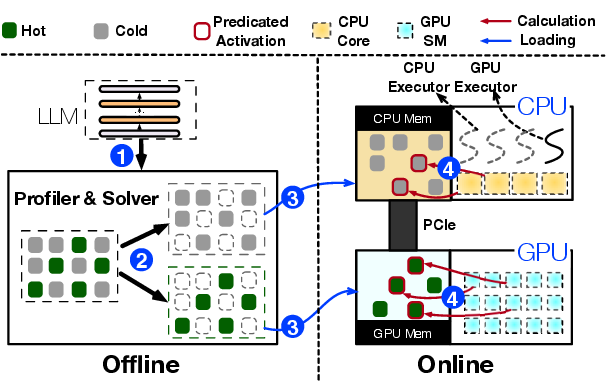}
    \end{minipage}
    \begin{minipage}{1\linewidth}
    \vspace{-5mm}
    \caption{The architecture and inference workflow of {\sys}.}
    \vspace{-5mm}
    \label{fig:arch}
    \end{minipage} 
\end{figure}

Figure~\ref{fig:arch} presents an architectural overview of {\sys},
comprising both offline and online components.
Due to the variation in locality properties among different LLMs,
the offline component should profile LLMs' activation sparsity,
differentiating between hot and cold neurons.
In the online phase, the inference engine loads two types of neurons into both GPU and CPU,
serving LLM requests with low latency during runtime.

\textbf{LLM Profiler and Policy Solver (Offline):}
Based on \textit{Insight-1}, the neuron's activation pattern can be collected enough from general datasets,
so we take an offline profiler that collects activation data from inference processes using requests derived from general datasets (e.g., C4~\cite{raffel2020exploring}).
It monitors neuron activation across all layers (Step \ding{172}),
followed by a policy solver categorizing neurons as hot or cold.
The solver aims to allocate frequently activated neurons to the GPU and others to the CPU.
It uses a neuron impact metric and hardware specifications to balance the workload,
using integer linear programming to maximize the GPU's impact metric for neurons (Step \ding{173}).

\textbf{Neuron-aware LLM Inference Engine (Online):}
Before processing user requests,
the online engine assigns the two types of neurons to their respective processing units (Step \ding{174}), as per the offline solver's output.
During runtime, the engine creates GPU and CPU executors, which are threads running on the CPU side, to manage concurrent CPU-GPU computations (Step \ding{175}).
The engine also predicts neuron activation and skips non-activated ones.
Activated neurons preloaded in GPU memory are processed there,
while the CPU calculates and transfers results for its neurons to the GPU for integration.
The engine uses sparse-neuron-aware operators on both CPU and GPU, focusing on individual neuron rows/columns within matrices.

\begin{figure}[t]
    \begin{minipage}{1\linewidth}
        \centering\includegraphics[scale=0.65]{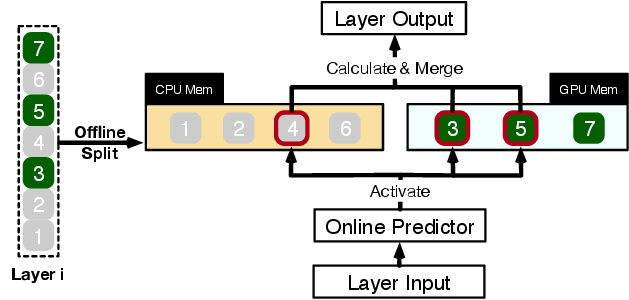}
    \end{minipage}
    \begin{minipage}{1\linewidth}
    \vspace{-4mm}
    \caption{An illustrative example shows how {\sys} calculates different neurons for one LLM layer.}
    \label{fig:example}
    \end{minipage} 
    \vspace{-5mm}
\end{figure}

\subsection{Single Layer Example}
Figure~\ref{fig:example} illustrates how {\sys} coordinates GPU and CPU in processing a layer's neurons.
It classifies neurons based on offline data, assigning hot-activated ones (e.g., indices 3, 5, 7) to GPU memory and others to CPU memory.
Upon receiving an input,
a predictor identifies which neurons in the current layer are likely to be activated.
For instance, it predicts activation for neurons 3, 4, and 5.
It is crucial to note that hot-activated neurons,
identified through offline statistical analysis,
may not consistently match the runtime activation behaviors.
For example, neuron 7, though labeled as hot-activated,
is forecasted to be inactive in this case.

Both the GPU and CPU concurrently process their respective predicted active neurons, while efficiently bypassing inactive ones.
Specifically, the GPU computes neurons 3 and 5, leveraging its parallel processing capabilities,
while the CPU simultaneously handles neuron 4.
Upon completion of neuron 4's computation on the CPU, its output is swiftly transferred to the GPU.
The GPU then performs a final integration step, combining the results from all activated neurons (3, 4, and 5) using an optimized add operator.
\vspace{-1mm}
\section{Neuron-aware Inference Engine}
\label{sec:design}
\subsection{Adaptive Sparsity Predictors}
\label{subsec:predictor}

The online inference engine
in {\sys} reduces computational loads by only processing those neurons that are predicted to be activated.
This method was also used in DejaVu~\cite{liu2023deja},
which advocates for training a set of fixed-size MLP predictors.
Within each Transformer layer, DejaVu utilizes two separate predictors to forecast the activation of neurons in the self-attention and MLP blocks.
Consequently, the inference computation is confined to neurons predicted to be active.

However, designing effective predictors for local deployments with limited resources is challenging,
balancing prediction accuracy and model size.
These predictors, frequently invoked for neuron activation prediction, should be stored in GPU memory for fast access.
Yet, the considerable memory requirements of numerous fixed-size predictors can encroach upon the space needed for storing LLM parameters.
For example, predictors for the OPT-175B model require around 27GB of GPU memory, surpassing an NVIDIA RTX 4090 GPU's capacity.
On the other hand,
naively reducing predictor size impairs accuracy; a decrease from 480MB to 320MB in predictor size dropped its accuracy from 92\% to 84\%,
further adversely affecting the overall LLM accuracy (e.g., winogrande~\cite{sakaguchi2021winogrande} task accuracy from 72.77\% to 67.96\%).

\begin{figure}[t]
    \begin{minipage}{1\linewidth}
        \centering\includegraphics[]{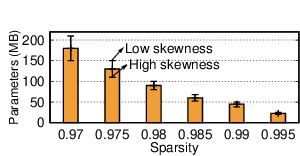}
    \end{minipage}
    \begin{minipage}{1\linewidth}
        \vspace{-4mm}
        \caption{\small{{Correlation between predictor parameter size and layer sparsity at a guaranteed 95\% accuracy level for OPT-175B.}} The X-axis represents sparsity, and the Y-axis represents the predictor parameter size. The bar indicates the average parameter size for the model in the corresponding sparsity, while the error bar reflects fluctuations in the predictor parameter size due to skewness within the layer.}
    \label{fig:design_predictor}
    \end{minipage} 
    \vspace{-5mm}
\end{figure}

We have observed that the size of predictors is influenced by two main factors: the sparsity of LLM layers and their internal skewness.
As shown in Figure~\ref{fig:design_predictor}, layers with higher activation sparsity simplify the task of identifying activated neurons,
allowing for smaller predictor models.
In contrast, layers with lower activation sparsity necessitate larger models with more parameters,
as accurately pinpointing activated neurons becomes increasingly challenging.
Additionally, in cases of high skewness, where activations are heavily concentrated in a few neurons,
even a compact predictor can achieve high accuracy.

To optimize for these factors,
We design an iterative training method for \textbf{non-fixed-size} predictors for each Transformer layer.
The process begins by establishing a baseline model size based on the layer's sparsity profile (Figure~\ref{fig:design_predictor}).
Subsequently, the model size is iteratively adjusted,
considering the internal activation skewness to maintain accuracy.
An MLP predictor typically comprises a input, a single hidden, and an output layers.
Since the dimensions of the input and output layers are determined by the Transformer layer's structure,
modifications primarily target the hidden layer.

In training predictors, model perplexity on the WikiText-2 dataset served as the benchmark for the predictive accuracy.
Initially, the predictor's dimensionality was set in accordance with the sparsity level of the layer.
We then engaged in an iterative process to tune the size of the hidden layer.
This process was continued until the perplexity of the model with the integrated predictor closely approximated that of the baseline model,
achieving a discrepancy of less than 0.1\%. 
Through this approach, {\sys} effectively limits predictor parameters to a mere 6\% of the total LLM parameters.












\subsection{Neuron Management}
\label{subsec:neuron_split}

When the offline solver determines a neuron placement policy,
the online inference engine of {\sys} loads the model into the CPU and GPU memory as per the policy.
For each layer, which may consist of multiple weight matrices,
{\sys} assigns each neuron to either the GPU or CPU based on whether the neuron is hot-activated.

Ensuring the accurate computation of these segmented neurons in their proper sequence is vital for precise results.
To this end,
{\sys} creates two neuron tables, one located in the CPU and the other in the GPU memory.
These tables correlate each neuron to its original position in the matrix.
During the process of multiplying with an input tensor,
each neuron interacts with its corresponding tensor value, guided by the mappings in the neuron tables.
The additional memory required for these neuron tables is relatively minimal,
totaling only about 9MB for an LLM like OPT-175B, which needs 350GB of storage.


\subsection{GPU-CPU Hybrid Execution}
\label{subsec:hybrid_exec}

{\sys} implements a GPU-CPU hybrid execution model where both units independently compute their respective activated neurons.
This approach balances the computational workload between GPU and CPU, leveraging the strengths of each unit while minimizing transfer time inefficiencies.
The GPU processes preloaded hot neurons, while the CPU handles cold neurons, with results combined on the GPU.

Before inference, {\sys} constructs a computationally directed acyclic graph (DAG)
with each node representing a computational LLM inference operator 
and stores it in a global queue in the CPU memory.
Each operator in the queue is tagged with its prerequisite operators.
During inference, two types of executors, pthreads created by the host OS,
manage calculations on both CPU and GPU.
They pull operators from the global queue, check dependencies,
and assign them to the appropriate processing unit.
The GPU and CPU use their neuron-aware operators,
with the GPU executor launching GPU operators using APIs like cudaLaunchKernel,
and the CPU executor coordinating unoccupied CPU cores for calculations.
Before executing an operator, the CPU executor also determines the necessary thread count for parallel computation.
To manage operator dependencies, especially when a parent node of a CPU operator is processed on the GPU,
a \textit{barrier} ensures GPU computations are complete before the CPU starts its operator.

In scenarios where activated neurons are split between GPU and CPU,
synchronization between these processing units also becomes crucial.
After one unit finishes its neuron calculations,
it waits for the other to merge results.
As GPU neurons are activated more frequently,
{\sys} assigns merging operations to the GPU.

\subsection{Neuron-aware Operator}
\label{subsec:operator}

Considering the activation sparsity in LLMs, matrix multiplication operations can bypass inactive neurons and their weights,
necessitating the use of sparse operators.
However, current sparse matrix multiplication tools,
including state-of-the-art sparse-aware compilers like SparTA~\cite{280848} and FlashLLM~\cite{xia2023flashllm},
as well as libraries like cuSPARSE~\cite{cuSPARSE},
fall short in this regard.
They either support only static compilation of sparse-aware kernels or require dynamic conversion of sparse matrices into dense formats,
leading to significant performance overhead,
especially with the dynamic sparsity in our scenario.
Additionally, the dynamic JIT compiler PIT~\cite{zheng2023pit},
though efficient for general sparse matrix multiplication on GPUs,
is not suited for CPU-GPU hybrid execution where CPU computational capabilities are limited.

To overcome these limitations,
{\sys} introduces neuron-aware operators
that directly compute activated neurons and their weights on both GPU and CPU without the need for runtime conversion to dense format.
These operators differ from traditional ones as they focus on individual row/column vectors within a matrix rather than the entire matrix.
They first determine a neuron's activation status and then process it if predicted to be active,
alongside the corresponding row or column of the parameter matrix.

\textbf{Neuron-aware Operators for GPU:}
Despite vector-vector calculations being less efficient than matrix-vector calculations on GPU,
neuron-aware operators based on vector-vector computation are advantageous when the batch size is small.
They avoid unnecessary computations and memory operations associated with inactive neurons and do not need costly matrix conversions.
Furthermore, these operators allow all thread blocks to concurrently check neuron activations and compute corresponding vectors if activated.
Notably, {\sys} assigns independent computational tasks to different thread blocks on the GPU.
Each thread block processes a distinct set of neurons, eliminating the need for synchronization or divergence across blocks.

\textbf{Neuron-aware Operators for CPU:}
Neuron-aware operators are particularly beneficial for CPUs,
which generally have lower parallelism.
The CPU executor assigns a neuron-aware operator to multiple cores,
dividing neurons into smaller batches for concurrent activation checking.
Each core processes only the activated neurons in its batch,
optimizing vector-vector calculations with hardware vector extensions like AVX2.
\vspace{-1mm}
\section{Neuron Placement Policy}
\label{sec:policy}
To fully harness the computational power of GPUs and CPUs, a locality-aware policy is crucial for optimizing performance.
If we randomly place neurons on GPU and CPU, the GPU cannot fully utilize its powerful computation capability because some hot neurons are placed on CPU.
Further, simply assigning the hotest neurons to the GPU may lead to excessive data transfers between the CPU and GPU, undermining efficiency.
To tackle this challenge, \sys{}'s offline component devises a placement policy using a solver that determines the allocation of each neuron to GPU or CPU.

\subsection{Offline Profiling}
Before determining the placement of each neuron, it is important to profile the activation information of each neuron.
Based on \textit{Insight-1} that the hot neurons in general corpus are also activated frequently across different scenarios, 
we achieve the profile with an offline profiler, which deploys the LLM to handle requests generated from multiple general datasets, such as C4~\cite{raffel2020exploring} and Wikipedia~\cite{wikidump}.
To accurately measure activation information, the profiler inserts a monitoring kernel after each block within a Transformer layer.
Additionally, it builds a neuron information table on the GPU, designed to track the activation count of each neuron.


\subsection{Neuron Impact Metric}
The neuron impact metric measures each neuron's contribution to the LLM's overall inference outcome, crucial for GPU neuron allocation.
We calculate this metric effectively by leveraging the fact that profiled activation frequency mirrors runtime behavior accurately, provided the profiling involves a substantial amount of input data.
As Equation \ref{eqn1} shows, this metric for a neuron is defined by its activation frequency obtained during profiling.

\vspace{-5mm} 
\setlength{\jot}{-5.0pt}
\normalsize{
\begin{gather*}
\label{eqn1} v_{i} = f_{i} \hspace{1cm}\forall i \in \mathbb{N} \tag{1}
\end{gather*}
}
\setlength{\jot}{0pt}
\vspace{-4mm}

\begin{table}
    \caption{Terminology for ILP formulation. The Par represents the parameters gathered from the profiler or the expressions used to define constraints, 
    none of which need to be solved by the solver. 
    The Var refers to the constraint and objective variables that emerge from the modeling process, which need to be solved by the solver.}
    \label{tab:terminology}
\scriptsize{
\centering
\begin{tabular}{clp{5cm}}
\hline
\textbf{Symbol} & \textbf{Type} & \textbf{Description} \\[1ex]
\hline
$\mathbb{L}$ &  Par & All layers \\
$\mathbb{N}$ & Par & All neurons \\
$\mathbb{U}$ & Par & CPU and GPU\\
$f_{i}$ &  Par & Activation frequency of neuron i \\
$N_{i}$ &  Par & Neuron in layer i \\
$v_{i}$ &  Par & Neuron impact for neuron i \\
$M_{i}$ & Par & The memory size for neuron i \\
$MCap_{j}$ &  Par & The memory size for processing unit j \\
$Bandwidth_{j}$ &  Par & The memory bandwidth for processing unit j \\
$T_{sync}$ &  Par & The time required for one synchronization between the CPU and GPU \\
$K$ & Par   & A large positive number \\
$a_{in}$ &  Var & Whether neuron n is placed on processing unit i \\
$T_{l}^{j}$ &  Var & The time for computing one neuron in layer $l$ on processing unit j \\
$C_{l}$ & Var & The minimum number of neurons required to be allocated on the GPU when the solver opts to split neurons in layer $l$ \\
$y_{l}$ &  Var & Binary auxliary variable for layer $l$ to facilitate the modeling of conditional constraints \\
\hline
\end{tabular}
}
\vspace{-2mm}

\end{table}

\subsection{Modeling of Neuron Placement}
Based on the neuron impact metric,
{\sys} utilizes a solver to optimize the total impacts of all neurons in the GPU. 
This cumulative impact is formulated as the objective function, as defined in Equation \ref{eqn2}.
This function is then input into an integer linear programming framework to identify a specific solution that maximizes the objective function.
The binary variable $a_{in}$, defined in Equation \ref{eqn3} indicates whether the neuron $n$ is placed on processing unit $i$.
\vspace{-2mm}
\setlength{\jot}{-0.4pt}
\normalsize{
\begin{gather*}
\label{eqn2} Maximize \quad t_i= \sum_{e \in \mathbb{N}} a_{ie} * v_{e} \forall i \in \{GPU\} \tag{2} \\
\label{eqn3} \sum_{i \in \mathbb{U}} a_{in} = 1 \quad\forall n \in \mathbb{N} \tag{3}
\end{gather*}
}
\setlength{\jot}{0pt}
\vspace{-2mm}

When maximizing the objective function, 
the solver also needs to consider two sets of constraints. First, minimize the \textbf{communication overhead} between processing units. 
Second, ensure the GPU memory is fully utilized.

\subsubsection{Communication Constraint}
The number of neurons preloaded onto the GPU is limited by layer communication overheads, constrained by hardware PCIe bandwidth. Preloading too few neurons negates the GPU's computational benefits. Consequently, the solver must determine a minimum neuron allocation for optimal GPU processing efficiency, as detailed in Inequality \ref{eqn4}.
In this inequality, $C_{l}$ is the minimum count of neurons that must be assigned to the GPU for layer $l$.

When solving Inequality \ref{eqn4},
it is essential to profile both the computation time for an individual neuron and the intra-layer communication overhead, $T_{sync}$.
In LLM inference, particularly with smaller batch sizes, 
the limiting factor is memory bandwidth. 
Hence, a neuron’s computation time is roughly equal to the duration required to access its weights once, as shown in Equation \ref{eqn5}.
And the extent of intra-layer data transfer tends to be consistent across layers,
leading to a uniform synchronization cost.
Consequently, we describe $T_{sync}$ as the profiled overhead for a single instance of intra-layer communication.

\vspace{-4mm}
\setlength{\jot}{-0.2pt}
\normalsize{
\begin{gather*}
\label{eqn4} C_{l} \cdot T_{l}^{GPU} + T_{sync} \leq C_{l} \cdot T_{l}^{CPU} \forall l \in \mathbb{L} \tag{4} \\
\label{eqn5}T_{i}^{j} = M_{i} / Bandwidth_{j} \quad \forall j \in \mathbb{U}, \forall i \in \mathbb{L} \tag{5} \\
\end{gather*}
}
\setlength{\jot}{0pt}
\vspace{-12mm}

\subsubsection{Memory Constraint}
Neuron placement is further constrained by the memory capacities of the processing units, as defined in Inequality \ref{eqn6}.
Moreover, the solver ensures that when allocating neurons of a layer to the GPU,
it either assigns at least the minimum number of neurons specified in Inequality \ref{eqn4} to offset communication costs
or opts not to allocate any neurons from that layer to the GPU.
Specifically, the number of neurons for layer $l$ on the GPU must either exceed $C_{l}$ or be equal to zero.

To model this constraint,
we introduce an auxiliary binary variable, $y_{l}$, which can be either 1 or 0.
This variable determines whether any neurons are assigned to the GPU for layer $l$.
For computational convenience, a sufficiently large number $K$ is also introduced.
Inequalities \ref{eqn7} and \ref{eqn8} are formulated to model this constraint.
When $y_{l}$ is 1, indicating neuron placement on the GPU for this layer, and given that $K$ is adequately large,
these two inequalities effectively become $C_l \leq \sum_{e \in N_l} a_{ie} \leq K$.
Conversely, if $y_{l}$ is set to 0,
signifying no neuron placement on the GPU for layer $l$, the inequalities reduce to $\sum_{e \in N_l} a_{ie} = 0$.

\vspace{-4mm}
\setlength{\jot}{-0.2pt}
\normalsize{
\begin{gather*}
\label{eqn6} \sum_{n \in N} a_{jn} \cdot M_{n} < MCap_j \quad \forall j \in \mathbb{U} \tag{6} \\
\label{eqn7}\sum_{e \in N_l} a_{ie} \geq C_l \cdot y_l \quad \forall l \in \mathbb{L}, \forall i \in \{GPU\} \tag{7} \\
\label{eqn8}\sum_{e \in N_l} a_{ie} \leq K \cdot y_l \quad \forall l \in \mathbb{L}, \forall i \in \{GPU\} \tag{8} 
\end{gather*}
}
\setlength{\jot}{0pt}
\vspace{-4mm}

\subsubsection{ILP Optimization}
The solver utilizes Integer Linear Programming (ILP) to maximize the objective function, conforming to all the constraints from Equation/Inequality \ref{eqn3} to \ref{eqn8}.
Given that ILP problems are inherently NP-complete,
directly solving them for an LLM with hundreds of billions of parameters poses a considerable computational challenge.
To expedite the process and achieve an approximate solution,
the primary strategy involves aggregating neurons within each layer into batches for collective placement analysis.
Specifically, the solver groups 64 neurons with similar impacts from a layer into a single batch.
This batching strategy dramatically reduces the total neuron count, N, from several millions to roughly tens of thousands,
thereby significantly decreasing the time to solve the ILP problem to approximately 10 seconds.

\vspace{-1mm}
\section{Implementation}
\label{sec:impl}

The online inference engine of {\sys} has been implemented
by incorporating 4,200 lines of C++ and CUDA code into llama.cpp~\cite{llama.cpp},
a state-of-the-art open-source LLM inference framework designed for PCs.
The extensions made by {\sys} include modifications to the model loader for distributing an LLM across GPU and CPU,
following the guidance from the offline solver's outputs.
We have also optimized the inference engine for GPU-CPU hybrid execution and introduced 10 neuron-aware operators for both processing units. 
All other components and functionalities of llama.cpp remains unchanged.
For instance, the KV cache continues to reside in CPU memory, allowing more GPU memory for hot-activated neurons.
Furthermore, around 400 lines of Python code were added to the transformers framework~\cite{wolf-etal-2020-transformers},
enabling it to function as an offline profiler and solver for {\sys}.

The current implementation of {\sys} supports a range of mainstream LLM families with varying parameter sizes.
For these models, {\sys} utilized general corpora such as Wikipedia~\cite{wikidump} to train online activation predictors,
Notably we deliberately avoided using any of the downstream task datasets in our predictor training process.
Training predictors, though time-consuming (often several hours), is a one-time task that can be expedited with multiple GPUs.
\vspace{3mm}
\section{Evaluation}
\label{sec:eval}

\subsection{Experimental Setup}

\textbf{Hardware.}
All experiments were conducted on two distinct PC configurations,
representing both high-end and low-end hardware scenarios:
\begin{itemize}
    \item \textbf{PC-High}: Intel i9-13900K processor (eight 5.4GHz cores), 192GB host memory (bandwidth of 67.2 GB/s),
    an NVIDIA RTX 4090 GPU (24GB), and PCIe 4.0 interface (64GB/s bandwidth).
    \item \textbf{PC-Low}: Intel i7-12700K processor (eight 4.9GHz cores), 64GB host memory (bandwidth 38.4 GB/s),
    an NVIDIA RTX 2080Ti GPU (11GB), and PCIe 3.0 interface (32GB/s bandwidth).
\end{itemize}

\noindent\textbf{Models.}
Table~\ref{tab:model_eval} shows the LLMs evaluated in this section, including average activation sparsity for various LLMs in the MLP blocks.
These LLMs include OPT~\cite{zhang2022opt} models with parameters from 13B to 175B, Bamboo 7B~\cite{bamboo},
as well as Falcon(ReLU)-40B~\cite{falcon40b} and LLaMA(ReGLU)-70B~\cite{LLaMA-70B} models.
In addition to LLMs using ReLU-family activation functions, 
we also evaluate SiLU-family LLMs, with sparsity levels of approximately 50\%.
For our experiments, all models in our experiments use FP16 and INT4 quantized parameters,
with intermediate activations in FP32, consistent with recent LLM research practices~\cite{yu2022orca,frantar-gptq}.


\begin{table}[!t]
    \caption{All LLMs used for evaluation and their average activation sparsity in the MLP blocks.
    The sparsity is measured using the same method in Table~\ref{tab:model_sparsity}.}
    \label{tab:model_eval}
    \vspace{-4mm}
    \resizebox{0.99\linewidth}{!}{
    \begin{tabular}{ccc}
    \toprule
    \textbf{Model} & \textbf{Activation Function} & \textbf{Sparsity} \\ \midrule
    Bamboo-7B & dReLU & 90\%  \\
    OPT-7B/13B/30B/66B/175B & ReLU & 96\%-98\%  \\
    Falcon(ReLU)-40B & ReLU & 95\%  \\
    LLaMA2(ReGLU)-7B & ReGLU & 70\%  \\
    LLaMA2(ReGLU)-13B & ReGLU & 78\%  \\
    LLaMA2(ReGLU)-70B & ReGLU & 82\%  \\
    Qwen1.5-4B & SwiGLU  & 40\%  \\
    LLaMA2(SwiGLU)-13B & SwiGLU& 43\%  \\
    Yi-34B & SwiGLU & 53\%  \\
    \toprule
    \end{tabular}
    }
\vspace{-4mm}
\end{table}

\noindent\textbf{Workloads.}
The workloads for our experiments are derived from chatbot arena~\cite{zheng2023judging} ChatGPT prompts~\cite{CP} and Alpaca~\cite{alpaca} datasets.
covering a wide spectrum of language model uses.
These datasets are the most representative examples of real LLM services.
ChatGPT prompts include user interactions with ChatGPT~\cite{ChatGPT},
and Alpaca features instruction sets generated by GPT3.5 through self-instruction.

\noindent\textbf{Baseline System.}
We compare {\sys} with llama.cpp~\cite{llama.cpp} and SpecInfer~\cite{miao2023specinfer}, state-of-the-art local LLM inference frameworks. 
llama.cpp is the most widely used LLM inference framework for local scenarios.
For a fair comparison, we extended llama.cpp to support the OPT model, as it does not natively do so. 
SpecInfer is representative speculative decoding framework,
that utilizes smaller draft models to generate tokens and subsequently verifies these tokens in batches using the original model.
While other alternatives like FlexGen~\cite{sheng2023flexgen} and DejaVu~\cite{liu2023deja} exist,
they exhibit higher latency in the latency-sensitive scenarios discussed in this paper,
as analyzed in $\S$\ref{subsec:bg-offloading}.

\subsection{End-to-End Performance}
\label{subsec:e2e}


\vspace{-2mm}
\begin{figure}[htp]
    \begin{minipage}{1\linewidth}
        \centering\includegraphics[width=0.9\linewidth]{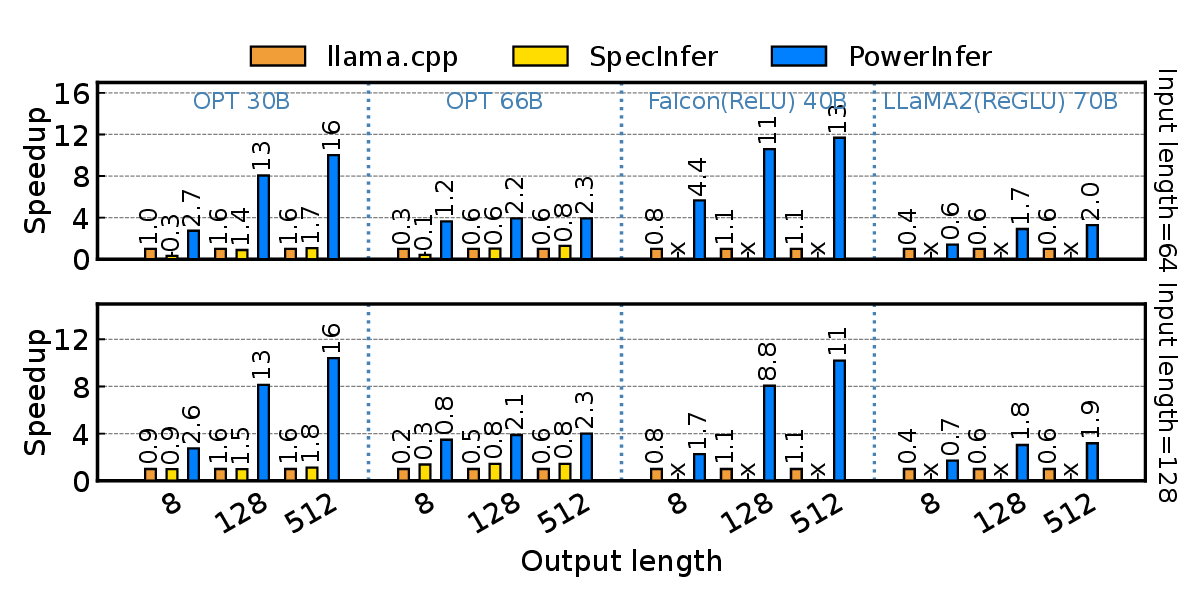}
    \end{minipage}
    \vspace{-5mm}
    \captionof{figure}{\emph{\small{Speedup of various models on PC-High in FP16 format.}}
    The X axis and Y axis indicate the output length and speedup compared with llama.cpp. 
    The number above each bar indicates the end-to-end generation speed (tokens/s).
    The up and down figures represent input length of 64 and 128.}
    \vspace{-5mm}
    \label{fig:eval-e2e-high}
\end{figure}
\begin{figure}[t]
    \begin{minipage}{1\linewidth}
        \centering\includegraphics[width=0.9\linewidth]{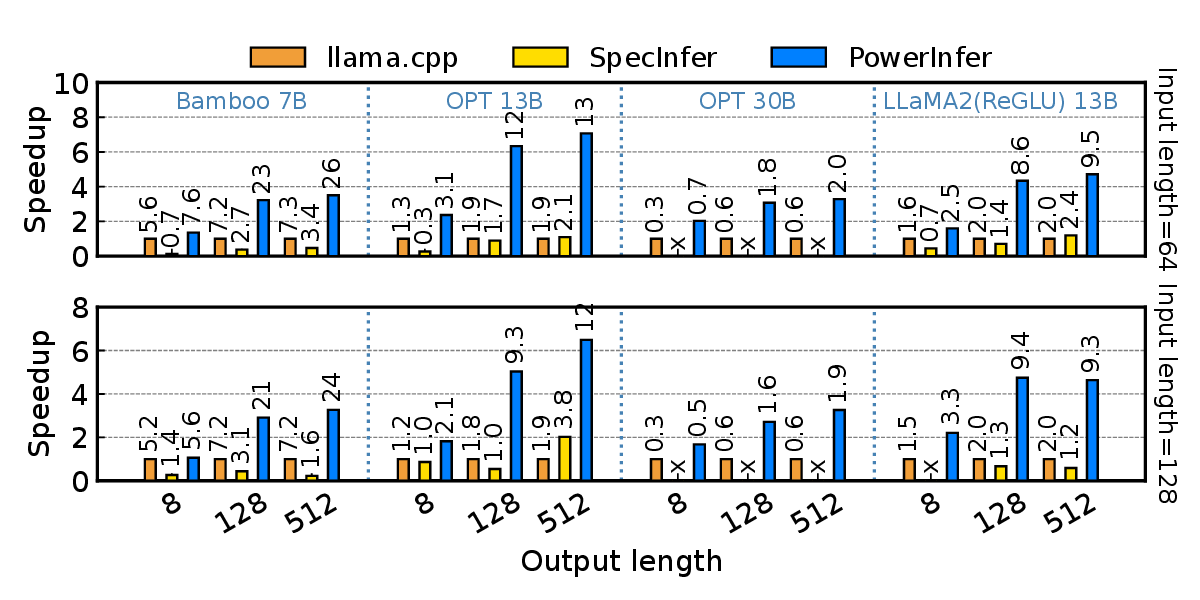}
    \end{minipage}
    \vspace{-5mm}
    \captionof{figure}{\emph{\small{Speedup of various models on PC-Low in FP16 format.}}
    The up and down figures represent input length of 64 and 128, respectively.}
    \vspace{-5mm}
    \label{fig:eval-e2e-low}
\end{figure}

We first compare the end-to-end inference performance of {\sys} , llama.cpp and SpecInfer with a batch size of one,
the typical setting for local deployments~\cite{medusa}.
Given real-world dialog input/output length variability~\cite{kwon2023efficient},
we sample prompts from Alpaca and ChatGPT datasets, ranging from 8 to 128 tokens and measure the generation speed.

Figure~\ref{fig:eval-e2e-high} illustrates the generation speeds for various models and input-output configurations on a PC-High equipped with an NVIDIA RTX 4090.
On average, {\sys} achieves a generation speed of 8.32 tokens/s, reaching up to 16.06 tokens/s,
significantly outperforming llama.cpp and SpecInfer with average speedups of 7.23$\times$ and 6.19$\times$, and for Falcon-40B, up to 11.69$\times$ compared with llama.cpp.
Although SpecInfer utilizes speculative decoding, since the large model size exceeds the GPU's capacity, there is still a significant amount of swapping during the verify phase, accounting for over 95\% of the time. As a result, SpecInfer does not significantly outperform llama.cpp.
The performance superiority of {\sys} becomes more pronounced as the number of output tokens increases 
since the generation phase plays a more significant role in the overall inference time.
In this phase, a small number of neurons are activated on both CPU and GPU,
leading to fewer unnecessary computations compared to llama.cpp.
For example, in the case of OPT-30B, only around 20\% of neurons are activated for each token generated,
with the majority processed on the GPU, a benefit of {\sys}'s neuron-aware inference engine.

Figure~\ref{fig:eval-e2e-low} shows that on a lower-end PC (PC-Low),
{\sys} still attains considerable performance enhancement over llama.cpp and SpecInfer,
averaging a speedup of 4.71$\times$, 5.97$\times$ and peaking at 7.06$\times$ and 7.47$\times$.
However, these improvements are smaller compared to those on a higher-end PC (PC-High),
primarily due to the 11GB GPU memory limitation of PC-Low.
This limitation affects the number of neurons that can be allocated to the GPU,
particularly for models with around 30B parameters or more,
leading to a greater dependence on the CPU for processing a larger number of activated neurons.

\begin{figure}[t]
    \vspace{-2mm}
    \subfloat[PC-High] {
        \includegraphics[width=0.47\linewidth]{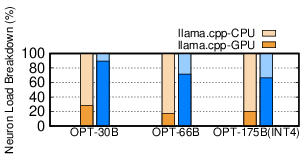}
        \label{fig:load-distribution-pc-high}
    }
    \subfloat[PC-Low] {
        \includegraphics[width=0.47\linewidth]{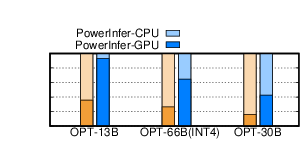}
        \label{fig:load-distribution-pc-low}
    }
    \vspace{-3mm}
    \caption{\small{\emph{Neuron load distribution on CPU and GPU during inference}}. The \textit{yellow} block refers to llama.cpp, and \textit{blue} block refers to {\sys}.}
    \vspace{-3mm}
    \label{fig:load-distritbution}
\end{figure}

Figure~\ref{fig:load-distritbution} presents the distribution of neuron loads between the CPU and GPU for both {\sys} and llama.cpp.
Neuron loads refer to the proportion of activated neuron computations carried out by each processing unit.
Notably, on PC-High, {\sys} significantly increases the GPU's share of neuron load, from an average of 20\% to 70\%.
This indicates that the GPU processes 70\% of activated neurons.
However, in cases where the model's memory requirements far exceed the GPU's capacity,
such as running a 60GB model on an 11GB 2080Ti GPU, the GPU's neuron load is reduced to 42\%.
This decrease is due to the GPU's limited memory, which is insufficient to host all hot-activated neurons,
necessitating that the CPU compute a portion of these neurons.

\noindent \textbf{Inference with different input length.}
In scenarios involving long input prompts with extremely short output lengths (less than 8 tokens),
which are less common~\cite{CP},
{\sys} demonstrates limited performance gains, ranging from 1.07$\times$ (Figure~\ref{fig:eval-e2e-low}) to 4$\times$ (Figure~\ref{fig:eval-e2e-high}).
In such situations, the prompt phase, where a substantial number of tokens are processed simultaneously,
becomes a crucial factor in determining inference speed.
This results in each token activating a unique set of neurons,
substantially diminishing activation sparsity.
As a consequence, the CPU becomes the primary bottleneck in the inference process,
tasked with processing a considerable number of cold-activated neurons but constrained by its computational capabilities.

For relatively long input sequences, {\sys} still achieves a 3.47$\times$ to 5.69$\times$ speedup, as shown in Table~\ref{tb:long}.
This is because when the input sequence is lengthy, all neurons are activated collectively by the input tokens. In this case, {\sys} switches to dense GPU computation during the prefill stage, resulting in prefill latency comparable to llama.cpp.
During the generation phase, where only one token is processed at a time, sparsity emerges in each inference step. Here, {\sys} leverages its hybrid engine to achieve significant acceleration.

\begin{table}[!t]
    \centering
    \caption{End-to-end latency comparison for different models with 1.5K input length and 256 output length on PC-High.}
    \label{tb:long}
    \vspace{-3mm}
    \resizebox{0.99\linewidth}{!}{
        \begin{tabular}{c|c|c|c}
            \hline
            LLMs & llama.cpp (ms) & PowerInfer (ms) & Speedup \\
            \hline
            LLaMA(ReGLU)-13B-FP16 & 49.91 & 14.38 & 3.47$\times$ \\
            Falcon(ReLU)-40B-FP16 & 321.63 & 56.48 & 5.69$\times$ \\
            LLaMA(ReGLU)-70B-FP16 & 92.76 & 37.17 & 2.50$\times$ \\
            \hline
            \end{tabular}
    }
    \vspace{-3mm}
\end{table}

\begin{figure}[t]
    \begin{minipage}{1\linewidth}
        \centering\includegraphics[width=0.9\linewidth]{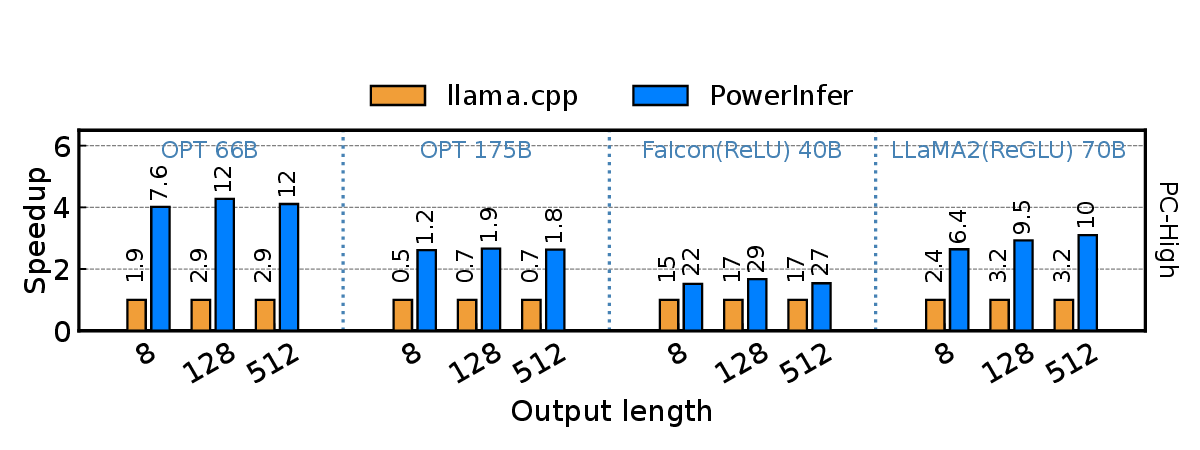}
    \end{minipage}
    \begin{minipage}{1\linewidth}
        \centering\includegraphics[width=0.9\linewidth]{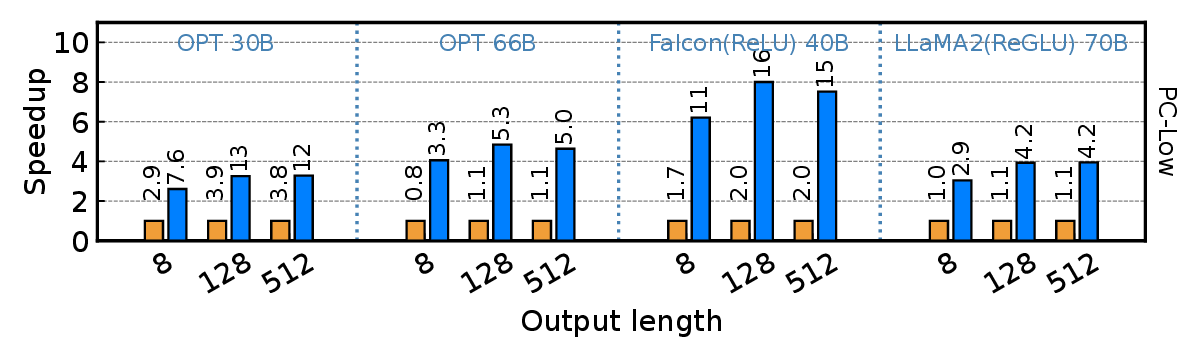}
    \end{minipage}
    \vspace{-5mm}
    \captionof{figure}{\emph{\small{Speedup of different models on PC-High and PC-Low in INT4 format.}}
    The upper row of the figure presents performance on PC-High, while the lower row details those on PC-Low.}
    \label{fig:int4}
    \vspace{-3mm}
\end{figure}

\noindent \textbf{Inference with Quantization.}
Figure~\ref{fig:int4} illustrates that {\sys} effectively supports LLMs that are compressed using INT4 quantization.
We fail to run SpecInfer with INT4 quantization due to the lack of support.
On a high-end PC (PC-High), {\sys} delivers responses at an average speed of 13.20 tokens/s,
reaching a peak of 29.08 tokens/s. 
The average speedup achieved compared with llama.cpp is 2.89$\times$, with a maximum of 4.28$\times$.
On a lower-end setup (PC-Low), the average speedup is 5.01$\times$, peaking at 8.00$\times$.
The reduction in memory requirements due to quantization enables {\sys} to
more efficiently manage larger models.
For instance, in our experiment with the OPT-175B model on PC-High,
{\sys} nearly reaches two tokens per second, surpassing llama.cpp by a factor of 2.66$\times$.

\begin{figure}[t]
    \begin{minipage}{1\linewidth}
        \centering\includegraphics[]{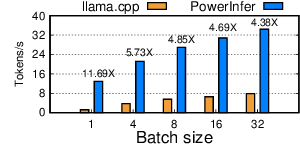}
    \end{minipage}
    \begin{minipage}{1\linewidth}
    \vspace{-3mm}
    \caption{\small{\emph{Batch inference speedup of Falcon-40B on PC-High.} The X axis indicates the request batch size, the Y axis represents the end-to-end token generation speed (tokens/s). The number above each bar shows the speedup compared with llama.cpp.}}
    \label{fig:batch}
    \end{minipage} 
    \vspace{-4mm}
\end{figure}

\noindent \textbf{Batching Inference.} 
We also evaluate the end-to-end inference performance of {\sys} with different batch sizes, as shown in Figure~\ref{fig:batch}.
{\sys} demonstrates a significant advantage when the batch size is smaller than 32, 
achieving an average 6.08$\times$ improvement in performance compared with llama.cpp.
As the batch size increases, the speed-up ratio offered by {\sys} decreases. 
This reduction is attributed to the diminished sparsity of model joint activations.
However, even with the batch size set to 32, {\sys} still maintains a considerable speedup, achieving a 4.38$\times$ speedup.

\subsection{Ablation Studies}
\label{subsec:ablation}
\subsubsection{Performance Breakdown}

\begin{figure}[t]
    \begin{minipage}{1\linewidth}
        \centering\includegraphics[]{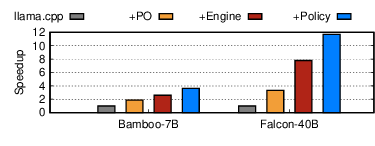}
    \end{minipage}
    \begin{minipage}{1\linewidth}
    \vspace{-5mm}
    \caption{\emph{\small{Performance breakdown for each component of {\sys}. The Falcon-40B is running on PC-High and Bamboo-7B is running on PC-Low.}}}
    \label{fig:ablation}
    \end{minipage} 
    \vspace{-6mm}
\end{figure}

Figure~\ref{fig:ablation} breaks down the contributions of each {\sys} component to the overall performance speedup.
Using a step-by-step integration method,
we progressively incorporate {\sys} features into llama.cpp.
First, we add {\sys}'s predictors and neuron-aware operators into llama.cpp (labeled "+PO"),
enabling computation of only activated neurons on both GPU and CPU.
Yet, +PO still adheres to layer-wise computation, where each layer is processed entirely by either GPU or CPU.

Building on +PO, we introduce {\sys}'s hybrid inference engine (denoted "+Engine"),
which allows neuron-aware operators to process neurons within the same layer simultaneously on both GPU and CPU.
+Engine uses a naive neuron partitioning policy that assigns neurons randomly to the GPU.
The final step involves integrating our optimized policy ("+Policy"),
formulated by the offline solver as described in $\S$\ref{sec:policy},
into the +Engine setup, showcasing the full capabilities of {\sys}.

The initial integration of +PO into llama.cpp yields performance boosts of 1.87$\times$ and 3.32$\times$ for Bamboo-7B and Falcon-40B, respectively,
primarily by reducing unnecessary inactive neurons.
+Engine further escalates these gains to 2.60$\times$ and 7.80$\times$,
thanks to precise neuron placement and intra-layer calculations that significantly increase the GPU's computational share.
Finally, incorporating +Policy results in improvements of 3.62$\times$ and 11.69$\times$.
The enhancement achieved by our policy lies in its ability to finely balance the intra-layer communication overhead.
The naive partitioning policy in +Engine overlooks the hotness of neurons and the GPU-CPU intra-layer communication,
often offsetting the benefits of assigning high-frequency activation neurons to the GPU.
Conversely, our policy in {\sys} more adeptly balances processing loads and communication costs between the CPU and GPU.

\subsubsection{Generation Latency Analysis}

In this section, 
we investigate the robustness of {\sys}'s speedup across different tasks and inter-token generation latency distritbution. 
We sample from the most representative tasks on the Hugging Face community, including STEM, coding, roleplaying and information extraction. 
We measure the inter-token generation latency distribution across different tasks. 
Table~\ref{tb:lat} presents the evaluation results of Bamboo-7B on PC-Low. 
\sys{} exhibits a highly consistent inter-token generation latency across different tasks, with only a 10\% difference between the P95 latency and the average. It is due to \sys{}'s policy that places the general hot neurons on GPU. 
This fluctuation is caused by the sparsity variation of different tokens from 80\% to 86\%. When encountering tokens with lower sparsity, the computation introduces 10\% more latency compared with the average.
\begin{table}[!t]
    \centering
    \caption{Generation latency (ms) distribution on various Tasks.}
    \label{tb:lat}
    \vspace{-4mm}
    \resizebox{0.99\linewidth}{!}{
    \begin{tabular}{lr|ccccc}
    \toprule
    & & STEM & HumanEval & Roleplay  & Table-Extraction\\
    \midrule
    \multirow{3}{*}{\parbox{1.75cm}{\raggedleft Bamboo-7B\\ on PC-Low}}
    & Avg. & 38.05 & 36.71 & 37.08 & 37.42\\
    & P95 & 40.75 & 40.48 & 40.61 & 40.74\\
    & P99 & 42.33 & 42.37 & 43.87 & 42.93\\
    \bottomrule
    \end{tabular}
    }
\end{table}

\subsubsection{Neuron-aware Operator Performance}

\begin{figure}[t]
\begin{minipage}{\linewidth}
\centering
\includegraphics[width=\linewidth]{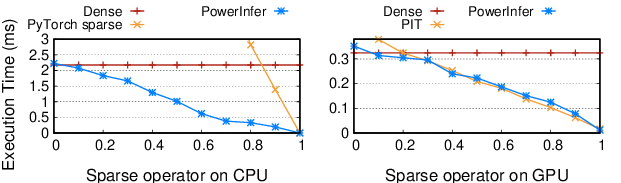}
\end{minipage} \\[2pt] %
\vspace{-3mm}
\begin{minipage}{1\linewidth}
\caption{\small{\emph{Comparing Neuron-aware operator with different sparse operators on PC-Low.} The X axis indicates the sparsity level, the Y axis represents the execution time (ms).}}
\label{fig:operator}
\end{minipage} 
\vspace{-5mm}
\end{figure}

This section evaluates the performance of {\sys}'s sparse operators on both CPU and GPU across various sparsity levels.
We benchmark {\sys} against leading sparse libraries: for CPU benchmarks, we use PyTorch sparse, the state-of-the-art sparse kernels within PyTorch, as our baseline.
In GPU, {\sys} is compared with PIT~\cite{zheng2023pit}.
Given that the sparsity in LLMs is typically based on neuron granularity,
our experiments are specifically designed to evaluate sparse matrices of this nature.
We focus on sparse matrix-vector multiplication using a [4096, 4096] $\times$ [4096, 1] configuration,
a common setup in local LLM inference~\cite{medusa}.
To adjust sparsity, we introduce zero values to matrix rows.

Figure~\ref{fig:operator} shows that {\sys}'s operator achieves nearly linear acceleration with increasing sparsity levels,
a stark contrast to dense matrix computations.
On the CPU, traditional sparse operators do not outperform dense computation until sparsity surpasses 87\%.
However, {\sys}'s CPU operator outperforms dense matrix multiplication even at sparsity levels below 10\%.
For the GPU, {\sys} matches PIT in performance.
Its primary advantage, however, is its unified CPU-GPU framework.
This design allows for flexible execution of sparse operators on both processing units, unlike PIT,
which is optimized solely for GPU-based sparse matrix multiplication and does not support hybrid CPU-GPU environments.

\subsubsection{Predictor Overhead}
\begin{figure}[t]
    \begin{minipage}{1\linewidth}
        \centering\includegraphics[]{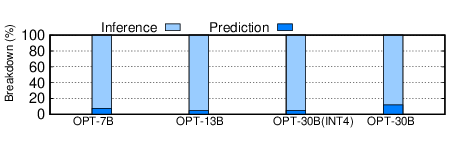}
    \end{minipage}
    \begin{minipage}{1\linewidth}
    \vspace{-8mm}
    \caption{\emph{\small{End-to-end prediction overhead of {\sys} on PC-Low.}} 
    The Y-axis displays the percentage breakdown between predictor overhead and LLM inference time.}
    \label{fig:predictor}
    \end{minipage} 
    \vspace{-5mm}
\end{figure}

\begin{table}[h]
    \caption{Predictor parameter sizes and ratios for various LLMs. The predictor ratio represents the percentage of predictor parameters relative to the original model parameters.}
    \label{tab:predictor_size}
    \vspace{-4mm}
    \centering
    \resizebox{\linewidth}{!}{
    \begin{tabular}{l|cccc}
    \hline
    Model  & OPT-13B & OPT-66B & Falcon(ReLU)-40B & LLaMA(ReGLU)-70B \\
    \hline
    Predictor-params &  0.88B  & 3.23B & 3.63B & 5.66B \\
    \hline
    Predictor-ratio(\%) & 6.71\% & 4.88\% & 8.68\% & 8.08\% \\
    \hline
    \end{tabular}
    }
\end{table}
The execution time of the online predictors for different models is also measured,
as depicted in Figure~\ref{fig:predictor}.
On average, the execution of predictors constitutes less than 10\% of the total inference time in {\sys}.
The execution time of predictors correlates directly with the predictor sizes,
which, as shown in Table~\ref{tab:predictor_size}, comprise only 7.09\% of the model weights.
The efficiency of these predictors stems from adaptive construction methods that minimize both size and computational load.
Furthermore, {\sys} integrates these predictors into its solver for neuron placement decisions, preferentially allocating them to GPUs.
This strategy exploits the parallel processing capabilities of GPUs, further reducing the runtime overhead associated with prediction.

\subsubsection{Performance Comparison with A100}
\begin{figure}[t]
    \vspace{-4mm}
    \begin{minipage}{1\linewidth}
        \centering\includegraphics[]{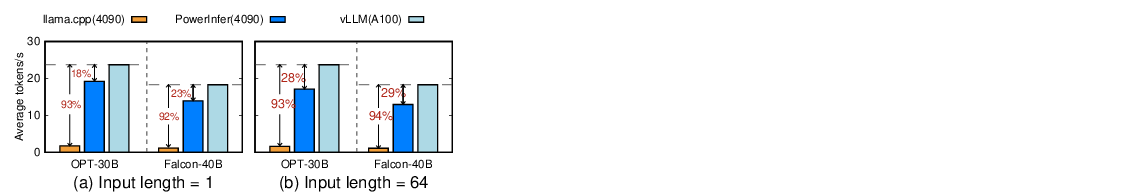}
    \end{minipage}
    \begin{minipage}{1\linewidth}
    \vspace{-2mm}
    \caption{\emph{\small{Generation speed of NVIDIA RTX 4090 compared with single A100.}} The X axis represents various models,
    while the Y axis represents generation speed (tokens/s) under various inference framework. 
    The percentages within the arrows represent the slowdown relative to vLLM on the A100.}
    \vspace{-2mm}
    \label{fig:A100}
    \end{minipage} 
    \vspace{-2mm}
\end{figure}

In our study, we analyze the extent to which {\sys} reduces the performance gap between a consumer-grade GPU and its top-tier server-grade counterpart.
Therefore, we evaluate the generation speed of {\sys}, deployed on PC-High,
in comparison to the performance of llama.cpp and vLLM~\cite{kwon2023efficient} executed on a single 80GB NVIDIA A100 GPU.
We chose the OPT-30B and Falcon-40B models for comparison, considering their exact memory requirements matching precisely with the capacity of the A100 GPU.
Our evaluation used input lengths of 1 and 64 to measure pure generation speed and conversational interactions, respectively.

Figure~\ref{fig:A100}a demonstrates that {\sys} significantly narrows the performance gap between the NVIDIA 4090 and A100 in generation tasks with input length 1.
On PC-High, llama.cpp lags behind vLLM on the A100 by 93\% and 92\% for OPT-30B and Falcon-40B, respectively, but {\sys} reduces this to 18\% and 23\%.
Figure~\ref{fig:A100}b shows that despite reduced cumulative sparsity in the prompt phase,
{\sys} still reduces the performance gap to 28\% and 29\%.
The remaining disparity mainly stems from the CPU's considerable computational load, which has become a bottleneck.

\begin{table}[!t]
\centering
\caption{Comparison of LLM accuracy between {\sys}-optimized models (termed as "model-{\sys}") and their original counterparts.
Arc-Challenge~\cite{arc} is a dataset for evaluating AI systems' comprehension and reasoning in natural language.
MMLU~\cite{hendryckstest2021} benchmarks model performance across various domains.
PIQA~\cite{bisk2020piqa} and Winogrande~\cite{sakaguchi2021winogrande}
assess common sense reasoning and understanding of physical interactions in LLMs.}
\label{tb:acc}
\vspace{-4mm}
\resizebox{0.99\linewidth}{!}{
\begin{tabular}{c|cccccc}
\toprule
& PIQA & Winogrande & Arc-Challenge & MMLU & GSM8K & Average\\
\midrule
\midrule
OPT-7B & 75.78\% & 65.19\% & 30.63\% & 24.95\% & 1.90\% & 39.69\%\\
OPT-7B-{\sys} & 75.67\% & 65.51\% & 30.63\% & 24.73\% & 1.82\% & 39.67\% \\
\midrule
OPT-13B & 76.01\% & 64.96\% & 32.94\% & 25.02\% & 2.12\% & 40.21\%\\
OPT-13B-{\sys} & 76.28\% & 65.98\% & 33.19\% & 24.76\% & 2.20\% & 40.48\%\\
\midrule
LLaMA(ReGLU)-13B & 76.44\% & 70.09\% & 36.52\% & 50.21\% & 25.40\% & 51.73\%\\
LLaMA(ReGLU)-13B-{\sys} & 74.06\% & 69.93\% & 36.60\% & 49.47\% & 23.90\% & 50.79\%\\
\midrule
Falcon-40B & 81.23\% & 75.45\% & 50.68\% & 51.78\% & 21.99\% & 56.23\%\\
Falcon-40B-{\sys} & 81.01\% & 75.92\% & 50.68\% & 51.68\% & 20.45\% & 55.95\%\\
\midrule
LLaMA(ReGLU))-70B & 82.01\% & 75.93\% & 52.39\% & 62.30\% & 62.30\% & 66.99\%\\
LLAMA(ReGLU)-70B-{\sys} & 82.05\% & 75.53\% & 51.45\% & 61.90\% & 61.90\% & 66.57\%\\
\bottomrule
\end{tabular}
\vspace{-5mm}
}

\end{table}

\subsection{SiLU-based LLM Performance}
\label{subsec:swiglu-performance}

\begin{table}[!t]
    \caption{Generation speed (tokens/s) comparison for SwiGLU and ReLU-based LLMs.}
    \label{tab:performance_silu}
    \vspace{-4mm}
    \resizebox{0.99\linewidth}{!}{
    \begin{tabular}{lc|ccc}
    \toprule
    \textbf{Setting} & \textbf{Model} & \textbf{PowerInfer} & \textbf{llama.cpp} & \textbf{Speedup} \\ \midrule
    PC-High & Yi(SwiGLU)-34B & 1.7 & 1.0 & 1.7$\times$ \\
    PC-Low & LLaMA(SwiGLU)-2-13B & 3.1 & 2.1 & 1.5$\times$ \\
    PC-High & OPT-30B & 12.0 & 1.1 & 10.9$\times$ \\ \bottomrule
    \end{tabular}
    }
    \vspace{-5mm}
\end{table}
While {\sys} demonstrates remarkable speedups on ReLU-based models due to their high activation sparsity,
it also shows effectiveness in accelerating SiLU-based models.
Our evaluation of SiLU-based LLMs, as presented in Table~\ref{tab:performance_silu},
reveals that {\sys} achieves a speedup of 1.47$\times$ to 1.7$\times$.
This performance gain, albeit less pronounced than for ReLU-based counterparts (like OPT-30B in Table~\ref{tab:performance_silu}),
underscores the efficacy of {\sys}'s sparse computation mechanisms across different activation functions.
The comparatively lower speedup can be attributed to the reduced sparsity in SiLU-based models,
where a larger proportion of neurons remain active during inference.
Consequently, CPU computation emerges as a potential bottleneck in these scenarios.
The speedup observed in SiLU models indicates that the effectiveness of {\sys}'s acceleration is correlated with the extent of the model's sparsity.

\subsection{LLM Accuracy}
\label{subsec:accuracy}

Since {\sys} selectively omits neurons predicted to be inactive,
we investigated whether this approach affects the inference accuracy of LLMs, focusing on models using ReLU or related activation functions.
Table~\ref{tb:acc} compares the accuracy of models from the OPT, Falcon (ReLU), and LLaMA2 (ReGLU) families,
both with and without differentiating activated/inactivated neurons, across a variety of representative downstream tasks.
The results show that {\sys} causes negligible loss in inference accuracy,
regardless of the model size or type of task,
consistent with previous research findings~\cite{liu2023deja}.
Although the predictors in each Transformer layer maintain an accuracy rate above 95\%,
they may occasionally miss some active neurons.
As a result, there are minor fluctuations in LLM accuracy,
leading to slight fluctuations in performance on specific downstream tasks.

We further analyze the nature and impact of mispredicted neurons on inference accuracy.
Our investigation reveals that neurons mispredicted by the predictor typically have minimal influence on the layer's output.
This characteristic makes these neurons more susceptible to misprediction, as their activation or non-activation has little effect on the overall result.
To quantify this impact, we conducted an analysis on the OPT-7B model.
By comparing the cosine similarity between outputs from original dense computations and our predictor-based computations,
we found that mispredictions only affect results by approximately 0.4\%.
This minimal difference further supports our observation that {\sys} maintains the accuracy of the original model,
despite the occasional misprediction of neurons.

\begin{table}[t]
    \caption{Performance comparison between SLM (Qwen-1.5-4B) and {\sys} (Bamboo-7B).}
    \label{tb:slm_comparison}
    \vspace{-4mm}
    \resizebox{\linewidth}{!}{
    \begin{tabular}{l|ccccc}
    \toprule
    Model & TBT(ms) & Average(\%) & MMLU(\%) & GSM8K(\%) & ARC-C(\%)  \\
    \midrule
    Qwen1.5-4B & 10.83 & 52.23 & 55.26 & 53.9 & 47.53  \\
    Bamboo-7B-{\sys} & 11.85 & 65.65 & 62.26 & 70.54 & 64.16  \\
    Bamboo-7B-dense & 18.54 & 65.49 & 62.46 & 70.28 & 63.74  \\
    \bottomrule
    \end{tabular}
    }
    \vspace{-5mm}
\end{table}
\textbf{Comparison with Smaller Language Models.}
To better understand {\sys}'s performance gains, we compare it with a common acceleration approach: 
using smaller language models (SLMs). However, the SLM approach often comes at the cost of reduced model accuracy. 
To fully understand the benefits of {\sys}, we compare its performance using a larger model 
to that of a state-of-the-art SLM, evaluating both speed and accuracy.
Table~\ref{tb:slm_comparison} shows that the Bamboo-7B model outperforms the Qwen-1.5-4B (a state-of-the-art 4B model)
across various tasks,
and {\sys} (Bamboo-7B-{\sys}) maintains the original dense model's accuracy
while achieving 4B-level decoding speed.
\section{Related Work}
\label{sec:related}

\textbf{LLM Weight Sparsity:}
Model pruning~\cite{Hansong,han2015deep,ma2023llmpruner} reduces parameters by setting weights to zero, as seen in SparseGPT~\cite{frantar-sparsegpt} and Wanda~\cite{sun2023wanda}, achieving 50\% sparsity. 
SparTA~\cite{280848} leverages both sparse tensor and SIMT cores by dividing sparse matrices.
Flash-LLM~\cite{xia2023flashllm} introduces a "Load-as-Sparse and Compute-as-Dense" approach for tensor core SpMM.
However, these methods, orthogonal to LLMs' intrinsic sparse activations,
usually incur accuracy losses and wall-clock model acceleration challenges~\cite{mirzadeh2023relu}.
In contrast, {\sys} uses natural sparse activations to maintain performance and efficiency.

\textbf{LLM Attention Sparsity:}
{\sys} leverages the sparse activation characteristics of MLP layers. It's worth noting that attention blocks also exhibit sparsity, which primarily stems from two sources:
attention heads, where only some need computation for the current token~\cite{liu2023deja}, and sparsity within heads, where unimportant KV cache is pruned or offloaded to alleviate memory-bound bottleneck of LLMs serving, as in H2o~\cite{zhang2024h2o} and InfiniGen~\cite{298683}.
The sparse activations used by {\sys} are orthogonal to attention sparsity and can further reduce inference latency.

\textbf{Speculative LLM Inference:}
Speculative inference~\cite{fu2023lookahead,medusa,tabi,chen2023accelerating} can also be leveraged to serve models exceeding GPU memory, 
which uses a smaller model to pre-decode tokens, later validated by the main model in a batch.
SpecInfer~\cite{miao2023specinfer} effectively reduces the number of LLM decoding steps.
While separate from our focus, integrating speculative inference into {\sys} could further boost LLM inference speed.

\textbf{LLM-Specific Serving Optimizations:}
There are many ML serving systems~\cite{clockwork} for general ML models. The prominence of Transformers has led to specialized serving systems~\cite{pets,fang2021turbotransformers,sheng2023slora, chen2023punica, khare2023superserve}.
Orca~\cite{yu2022orca} introduces iteration-level scheduling.
vLLM~\cite{kwon2023efficient} implements PagedAttention for token storage in varied GPU memory addresses,
overcoming KV cache's continuous storage limit.
These methods do not address the challenge of deploying models on PCs where the entire model cannot fit within the GPU memory.
\section{Conclusion}
\label{sec:concl}

{\sys} is a fast inference system optimized for LLMs that exploits the locality property in LLM inference.
It utilizes adaptive predictors and neuron-aware operators for neuron activation and computational sparsity.
{\sys} achieves up to 11.69$\times$ faster LLM inference compared to systems like llama.cpp, without compromising accuracy.

\section{Acknowledgments}
\noindent We sincerely thank our shepherd Shivaram Venkataraman and anonymous reviewers for their insightful suggestions.
We are deeply grateful to Rong Chen and Yubin Xia for their comprehensive and constructive feedback,
which greatly enhanced the quality of this paper.
This work was partially supported by National Key R\&D Program of China (2023YFB4503702) and NSFC (No. 62372287 and 61925206). 
Zeyu Mi (yzmizeyu@sjtu.edu.cn) is the corresponding author.

\bibliographystyle{ACM-Reference-Format}
\bibliography{ms}

\end{document}